\documentclass[11pt]{article}

\usepackage[letterpaper,margin=1in]{geometry}
\usepackage{amsmath}
\usepackage[round,authoryear]{natbib}

\let\href\undefined 
\usepackage[hidelinks]{hyperref}
\usepackage{layout}
\usepackage[capitalize]{cleveref}
\usepackage{xspace}
\usepackage{amsmath}
\usepackage{amssymb}
\usepackage{latexsym}  
\usepackage{xcolor}
\usepackage{booktabs}   
\usepackage{colortbl}
\usepackage{placeins}   
\usepackage{stfloats}   
\usepackage{wrapfig}    
\usepackage{capt-of}    
\usepackage{dsfont} 
\usepackage{float}  
\usepackage{ragged2e}
\usepackage[disable]{todonotes}
\usepackage{mathtools}   
\usepackage{graphicx}
\usepackage{epigraph}
\usepackage{siunitx}  
\usepackage{svg}
\usepackage{tikz}
\usetikzlibrary{tikzmark,decorations.pathreplacing,calc}
\usepackage{multirow}   
\usepackage{numprint}   
\npdecimalsign{.}
\usepackage{thmtools}
\usepackage{thm-restate}

\newenvironment{methods}{}{}

\title{PepALD: Macrocyclic Peptide Generation via\\ Autoregressive Latent Diffusion}

\author{
Junming Zhang\textsuperscript{1},
Siyu Yi\textsuperscript{2},
Wei Ju\textsuperscript{3},
and Zhonghui Gu\textsuperscript{4}\\[0.5em]
\small \textsuperscript{1}College of Computer Science, Sichuan University, Chengdu, 610065, China\\
\small \textsuperscript{2}School of Mathematics, Sichuan University, Chengdu, 610065, China\\
\small \textsuperscript{3}School of Artificial Intelligence, Sichuan University, Chengdu, 610065, China\\
\small \textsuperscript{4}Lingang Laboratory, Shanghai, 200031, China
}

\date{}

\begin{document}

\maketitle

\begin{abstract}
Macrocyclic peptides are promising therapeutic candidates for intracellular targets, but their design requires simultaneous control over non-natural monomer chemistry, ring topology, membrane permeability, and target binding.
Existing SMILES- or HELM-string generative models either operate in long atom-level sequence spaces or treat monomers as symbolic tokens with limited chemical grounding.
We introduce PepALD, an Autoregressive Latent Diffusion (ALD) foundation model for \textit{de novo} macrocyclic peptide generation.
The model represents HELM monomers with structured chemical embeddings, generates each residue through context-conditioned diffusion in chemically informed latent space, predicts R-group-aware ring closures during autoregressive generation, and aligns the denoiser to affinity rewards using winner-protected diffusion-adapted preference optimization.
In silico experiments demonstrate PepALD's generation quality and reward-optimization performance against representative peptide generation baselines.
\end{abstract}

\section{Introduction}\label{sec:introduction}

Conventional therapeutic modalities remain largely ineffective against many disease-relevant proteins, particularly those mediating protein-protein interactions (PPIs). Small molecules are inherently restricted to targeting deep, well-defined binding pockets and cannot effectively engage the large, shallow, and dynamic interfaces characteristic of PPIs \citep{bojadzic2018toward}. Monoclonal antibodies, while capable of high-affinity binding to these extended surfaces, suffer from obligate parenteral administration and are unable to access intracellular targets.
Macrocyclic peptides occupy a unique therapeutic niche. Their conformationally constrained backbones present preorganized interaction surfaces that match or exceed the footprint of antibody complementarity-determining regions (CDRs), enabling antibody-like potency and selectivity \citep{zorzi2017cyclic}. Critically, rational chemical optimization can confer clinically meaningful oral bioavailability, as demonstrated by the late-stage oral PCSK9 inhibitor enlicitide (MK-0616) \citep{gare2025lead}.

Despite remarkable advances in generating structures of target-binding cyclic peptides \citep{rettie2025accurate,rettie2025cyclic}, existing structural models are predominantly trained on natural amino acid proteins and peptides, limiting their ability to design cyclic peptides incorporating non-natural amino acids. Non-natural amino acids are indispensable for enhancing proteolytic stability, augmenting binding affinity via novel intermolecular interactions, and providing versatile attachment points for intramolecular cyclization \citep{hickey2023beyond}. Therefore, the development of generalizable foundation models for cyclic peptides that can accommodate non-natural amino acids and diverse cyclization chemistries is of paramount importance.


The simplified molecular-input line-entry system (SMILES) remains a dominant representation for molecular machine learning because it encodes atoms and bonds in a compact one-dimensional string \citep{ref_smiles}. Recent peptide generation models have therefore explored SMILES-space modeling. Geylan et al. present PepINVENT \citep{geylan2025pepinvent}, a transformer-based generative AI framework for SMILES generation, which enables de novo peptide design with both natural and non-natural amino acids and supports diverse topologies ranging from linear peptides to multiple macrocyclic architectures. Trained on a text-infilling task and coupled with reinforcement learning, PepINVENT facilitates multi-parameter optimization of peptides for therapeutic properties such as solubility, permeability, and topological constraints. Tang et al. introduce PepTune \citep{ref_peptune}, which formulates therapeutic peptide design as a multi-objective guided discrete diffusion process over peptide SMILES. While SMILES can precisely represent complex macrocyclic molecules and is broadly compatible with existing molecular property predictors, it encounters significant scalability challenges when applied to macrocyclic peptides. A single residue-level design decision may correspond to many coordinated SMILES tokens describing backbone atoms, side-chain atoms, stereochemistry, branches, and ring closures. Because available macrocyclic peptide datasets are far smaller than the resulting character-level state space, atom-by-atom generation can make validity learning data-hungry and can complicate reinforcement learning or guided diffusion, where sparse property rewards must be propagated across long token sequences.

Residue-level notation is therefore a natural bridge from molecular strings to peptide design, and HELM provides a standard abstraction. The hierarchical editing language for macromolecules is widely used for peptide and biopolymer therapeutics, including macrocyclic peptides, in databases and registration workflows \citep{ref_helm,ref_chembl}. HELM defines reusable monomers, polymer order, and inter-monomer bonds through explicit attachment points, encoding natural residues, non-natural residues, modifications, and cyclizations as monomer symbols plus typed R-group connections. This modular grammar suits macromolecules by retaining sequence compactness and covalent topology, while matching residue-level resources such as ChEMBL and CycPeptMPDB for data-driven macrocyclic peptide modeling \citep{ref_chembl,ref_cycpeptmpdb}.

HELM-GPT, a GPT-style decoder, has established HELM as a learnable language for \textit{de novo} macrocyclic peptide design and property optimization \citep{ref_helm_gpt}. However, a purely string-level HELM language model still treats monomer symbols largely as textual tokens, making its representations dependent on the limited peptide-sequence corpora available in this domain rather than on the underlying molecular structures of the monomers. Consequently, the explored chemical space is not guaranteed to extend beyond neighborhoods of the corpus distribution, which can leave rare building blocks under-explored, even when their physicochemical profiles closely resemble those of common monomers.

In this study, we propose PepALD, an AutoRegressive Latent Diffusion framework for \textit{de novo} macrocyclic peptide generation. PepALD adopts the HELM tokenization strategy, but embeds each monomer with structured representations derived from the Uni-Mol chemical foundation model \citep{ref_unimol}, so that generation is performed in a chemically informed latent space rather than over symbolic tokens alone. At each autoregressive step, a causal context encoder summarizes the generated prefix, a context-conditioned diffusion module samples the next monomer embedding, and a constrained mapper projects it back to an admissible HELM building block \citep{ref_ddpm,ref_ddim}. In parallel, an R-group-aware ring bond predictor reuses the autoregressive context together with pretrained attachment-site embeddings to determine whether, where, and how the peptide should cyclize. To further steer PepALD toward desired properties, we build on diffusion-adapted direct preference optimization \citep{ref_dpo,ref_diffusion_dpo} and introduce a winner-protected Diffusion-DPO (WP-DPO) objective inspired by DPO-Positive \citep{ref_dpop}, while constructing preference pairs with a structure-aware strategy that combines maximal marginal relevance diversity selection with nearest hard-negative coupling \citep{ref_mmr}.

We conduct extensive experiments comparing PepALD with representative baselines, including HELM-GPT and PepTune \citep{ref_helm_gpt,ref_peptune}. The model is first pretrained on ChEMBL32 HELM peptide sequences to learn general peptide syntax and monomer usage patterns, and then fine-tuned on CycPeptMPDB to capture macrocyclic peptide chemistry and topology \citep{ref_chembl,ref_cycpeptmpdb}. The resulting ALD model exhibits state-of-the-art generation performance, with leading validity, uniqueness, diversity, and novelty, as well as a balanced nearest-neighbor similarity profile. As target-focused case studies, we design cyclic peptide inhibitors against two protein targets: the intracellular SPSB2--iNOS protein--protein interaction \citep{sadek2018cyclic} and Mycobacterium tuberculosis chorismate mutase (MtbCM) \citep{van2025active}, using Uni-Dock/Vina docking scores as reward components \citep{ref_unidock}. Under the WP-DPO framework, the reward improves both the permeability-oriented and affinity-oriented objectives, while the solubility score is maintained. Equipped with chemically grounded latent representations, ring-closure prediction and WP-DPO reinforcement learning, PepALD exhibits extensive applicability in \textit{de novo} cyclic peptide design.

\begin{methods}
\section{Materials and Methods}\label{sec:methods}

\subsection{Overview}\label{subsec:overview}

We use HELM sequences mapped into the Uni-Mol latent space as model inputs. The model consists of three core components: a causal context encoder, a context-conditioned denoising network, and a ring-bond predictor. We then optimize the framework for target properties through a diffusion-adapted preference optimization procedure. An overview is shown in Figure~\ref{fig:architecture}.

\subsection{Representing HELM Grammar with Chemically Informed Monomer Embeddings}\label{subsec:helm}

We adopt HELM as the native sequence representation \citep{ref_helm}. Under the HELM specification, each monomer exposes a set of canonical attachment points---such as $\mathrm{R1}$, $\mathrm{R2}$, and $\mathrm{R3}$---through which all covalent bonds, backbone amides and intramolecular ring closures alike, are explicitly specified.

We next associate each monomer in the HELM vocabulary with a pre-computed embedding derived from Uni-Mol, a molecular representation model pretrained to capture three-dimensional molecular geometry and chemical semantics \citep{ref_unimol}. To preserve both residue-level identity and attachment-site chemistry, we extract a molecule-level representation for the whole monomer and atom-level representations for the atoms adjacent to the HELM R-group attachment sites. Missing R-group sites are represented by zero vectors.
Specifically, for each monomer $m$ with a SMILES representation, we extract a structured embedding vector:
\begin{equation}
\mathbf{e}_m = \bigl[\mathbf{e}_m^{\mathrm{CLS}},\; \mathbf{e}_m^{\mathrm{R1}},\; \mathbf{e}_m^{\mathrm{R2}},\; \mathbf{e}_m^{\mathrm{R3}}\bigr] \in \mathbb{R}^{4 \times d}
\label{eq:unimol_full}
\end{equation}
where $\mathbf{e}_m^{\mathrm{CLS}} \in \mathbb{R}^d$ is the molecule-level representation, and $\mathbf{e}_m^{\mathrm{R}i} \in \mathbb{R}^d$ ($i \in \{1,2,3\}$) are the atomic representations at the respective R-group attachment sites.

The embedding vector is pre-computed for all $|\mathcal{V}|$ monomers in the vocabulary and frozen during training, ensuring that the diffusion process operates in a chemically grounded latent space rather than a learned but potentially arbitrary token space.

\subsection{ALD generative model}\label{subsec:ald_generative_model}

Conditioned on a target length $T$, PepALD defines the probability of a HELM monomer sequence $\mathbf{x}=(x_1,\ldots,x_T)$ by an Autoregressive Latent Diffusion (ALD) process. At position $t$, the already generated prefix is summarized as a causal context vector $\mathbf{h}_t=f_{\mathrm{ctx}}(x_{<t})$, with a learnable start context used for $t=1$. The clean latent variable for the next residue is identified with the frozen Uni-Mol embedding of the corresponding monomer, $\mathbf{z}_{t,0}=\tilde{\mathbf{e}}_{x_t}$. Thus,
\begin{equation}
p_\theta(\mathbf{x}\mid T)=\prod_{t=1}^{T}p_\theta(x_t\mid x_{<t}),
\label{eq:ald_ar_factorization}
\end{equation}
where each discrete conditional distribution is induced by a continuous conditional density $p_\theta(\mathbf{z}_{t,0}\mid \mathbf{h}_t)$ followed by a chemically constrained projection to the HELM vocabulary:
\begin{equation}
p_\theta(x_t=m\mid x_{<t})=
\int \mathds{1}\!\left[\mathcal{M}_t(\mathbf{z},\mathbf{h}_t)=m\right]\,
p_\theta(\mathbf{z}\mid \mathbf{h}_t)\,d\mathbf{z}.
\label{eq:ald_induced_token_distribution}
\end{equation}
Here, $\mathcal{M}_t$ is the attachment-site-constrained token mapping rule that matches the closest monomer $m$ from the library.

\subsubsection{Forward diffusion}

For each valid residue position, the forward process perturbs the clean monomer embedding independently in the Uni-Mol latent space. Let $k\in\{1,\ldots,K\}$ denote the diffusion timestep, with variance schedule $\{\beta_k\}_{k=1}^{K}$, $\alpha_k=1-\beta_k$, and $\bar{\alpha}_k=\prod_{s=1}^{k}\alpha_s$. PepALD uses the standard DDPM Gaussian kernel \citep{ref_ddpm}:
\begin{equation}
q(\mathbf{z}_{t,k}\mid \mathbf{z}_{t,0},\mathbf{h}_t)
=\mathcal{N}\!\left(\sqrt{\bar{\alpha}_k}\mathbf{z}_{t,0},
(1-\bar{\alpha}_k)\mathbf{I}\right).
\label{eq:ald_forward_kernel}
\end{equation}
In the forward process, the noising kernel is conditionally independent of the autoregressive context, i.e., $q(\mathbf{z}_{t,k}\mid \mathbf{z}_{t,0},\mathbf{h}_t)=q(\mathbf{z}_{t,k}\mid \mathbf{z}_{t,0})$.
Equivalently,
\begin{equation}
\mathbf{z}_{t,k}=\sqrt{\bar{\alpha}_k}\mathbf{z}_{t,0}
+\sqrt{1-\bar{\alpha}_k}\boldsymbol{\epsilon},\quad
\boldsymbol{\epsilon}\sim\mathcal{N}(\mathbf{0},\mathbf{I}).
\label{eq:ald_forward_sample}
\end{equation}
During training, $k$ is sampled uniformly for each non-padding token, and the denoising network is trained to predict the injected noise $\boldsymbol{\epsilon}$ from $(\mathbf{z}_{t,k},k,\mathbf{h}_t)$.

\subsubsection{Reverse denoising with hybrid sampling}\label{subsubsec:reverse_hybrid_sampling}

Generation starts from isotropic Gaussian noise $\mathbf{z}_{t,K}\sim\mathcal{N}(\mathbf{0},\mathbf{I})$ for each autoregressive position. The reverse transition is parameterized by the context-conditioned noise predictor $\boldsymbol{\epsilon}_\theta$:
\begin{equation}
p_\theta(\mathbf{z}_{t,k-1}\mid \mathbf{z}_{t,k},\mathbf{h}_t)
=\mathcal{N}\!\left(\boldsymbol{\mu}_\theta(\mathbf{z}_{t,k},k,\mathbf{h}_t),
\sigma_k^2\mathbf{I}\right),
\label{eq:ald_reverse_kernel}
\end{equation}
with
\begin{equation}
\boldsymbol{\mu}_\theta=
\frac{1}{\sqrt{\alpha_k}}\left(
\mathbf{z}_{t,k}
-\frac{1-\alpha_k}{\sqrt{1-\bar{\alpha}_k}}\,
\boldsymbol{\epsilon}_\theta(\mathbf{z}_{t,k},k,\mathbf{h}_t)
\right),
\label{eq:ald_reverse_mean}
\end{equation}
and $\sigma_k^2=\beta_k(1-\bar{\alpha}_{k-1})/(1-\bar{\alpha}_k)$, with the stochastic term omitted at $k=1$ \citep{ref_ddpm}. For faster inference, the same predicted noise can be used in a DDIM trajectory over a reduced and adjustable subset of timesteps \citep{ref_ddim}.

After the final denoising step, the continuous sample is projected back to a valid HELM monomer using a hybrid decoder. At the core of the decoder is a fused score that reconciles the diffusion process with the contextual prior of the autoregressive state. Let $\tilde{\mathbf{e}}_m$ denote the $\ell_2$-normalized reference embedding of monomer $m$. For every admissible candidate $m \in \mathcal{C}_t$, we define:
\begin{equation}
\mathcal{S}(m \mid \mathbf{z}_{t,0}, \mathbf{h}_t)
= d(\mathbf{z}_{t,0}, \tilde{\mathbf{e}}_m)
- \lambda \log p_{\mathrm{LM}}(m \mid \mathbf{h}_t),
\label{eq:ald_hybrid_sampling}
\end{equation}
where $d(\cdot,\cdot)$ is the cosine distance in the shared embedding space, $p_{\mathrm{LM}}(m \mid \mathbf{h}_t)=\mathrm{softmax}(\mathbf{W}_{\mathrm{LM}}\mathbf{h}_t)_m$ is the categorical distribution produced by an auxiliary language-model head trained jointly with the denoiser, and the fusion weight $\lambda \geq 0$ is manually adjusted during sequence generation.
Related experiments are provided in Supplementary Section~10.

\subsection{Model architecture}\label{subsec:ald}

As shown in Fig.~\ref{fig:architecture}, PepALD implements the ALD formulation in Section~\ref{subsec:ald_generative_model} with three coordinated modules: context encoder, denoising network and ring-bond predictor.

\begin{figure}[H]%
\centering
\includegraphics[width=0.9\textwidth]{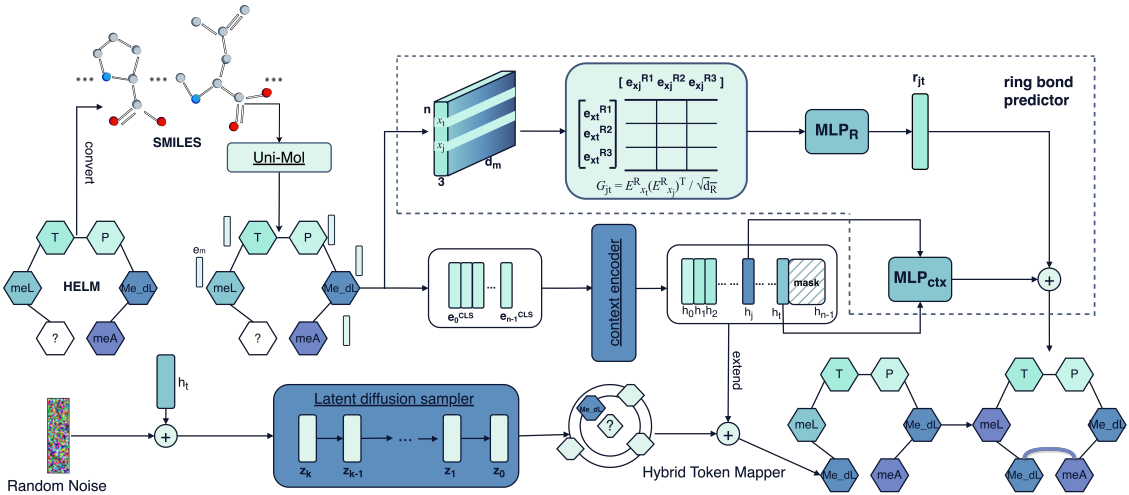}
\caption{Overview of the PepALD framework.
At each autoregressive step $t$, the Causal Context Encoder transforms the previously generated monomer embeddings $(\mathbf{e}_0, \ldots, \mathbf{e}_{t-1})$ into a context vector $\mathbf{h}_t$.
A context-conditioned Diffusion Engine then samples a continuous embedding $\mathbf{z}_t$ via iterative denoising in the Uni-Mol latent space.
The Token Mapper discretizes $\mathbf{z}_t$ into a HELM monomer token $x_t$ subject to position-dependent R-group chemical constraints.
Concurrently, the Autoregressive Ring Predictor evaluates potential ring-closure bonds between $x_t$ and all preceding residues using R-group compatibility scoring.
The process repeats until the target sequence length is reached, yielding a complete macrocyclic peptide with predicted ring topology.
}\label{fig:architecture}
\end{figure}

\subsubsection{Causal context encoder}\label{subsubsec:context_encoder}

To encode the previously generated monomers, we employ a causal context encoder $f_{\mathrm{ctx}}$.
Given the Uni-Mol embeddings of the generated prefix $(\tilde{\mathbf{e}}_{x_1}, \ldots, \tilde{\mathbf{e}}_{x_{t-1}})$, the encoder outputs a position-specific context vector $\mathbf{h}_t$ that conditions the diffusion model for the next monomer.

The encoder takes the prefix Uni-Mol embeddings with added sinusoidal positional encodings as input.
For efficient teacher-forced training, the embedded HELM sequence is processed in parallel by $L_{\mathrm{ctx}}$ causal Transformer layers \citep{ref_transformer}.
Within this stack, each causal Transformer layer consists of masked multi-head self-attention, a feed-forward block, residual connections, layer normalization, and dropout.
Through the causal mask, the hidden state at position $i$ can depend only on residues $x_{\leq i}$ in the sequence.

The output of the stack is then passed through an additional final layer normalization and a linear context projection, giving the overall context encoder flow:
\begin{equation}
\mathbf{H} = \mathrm{Proj}_{\mathrm{ctx}}\!\left(\mathrm{LN}\!\left[\mathrm{CausalTransformer}\!\left(\mathrm{PosEnc}(\mathbf{W}_{\mathrm{in}}\tilde{\mathbf{E}})\right)\right]\right).
\label{eq:causal_encoder}
\end{equation}
Here $\tilde{\mathbf{E}}$ denotes the matrix of input Uni-Mol monomer embeddings, $\mathbf{H}=[\bar{\mathbf{h}}_1,\ldots,\bar{\mathbf{h}}_T]$ denotes the sequence of contextual hidden states, and $\mathbf{W}_{\mathrm{in}}$ is the input projection to the Transformer hidden dimension.
For the first residue, where no prefix is available, we introduce a learnable start embedding and encode it as $\mathbf{h}_{\mathrm{start}}$.
For subsequent positions, the context used to predict residue $x_t$ is defined as the previous encoder hidden state, $\mathbf{h}_t=\bar{\mathbf{h}}_{t-1}$.

\subsubsection{Context-conditioned denoising network}

Given the context vector $\mathbf{h}_t$, we introduce a context-conditioned denoiser that guides the reverse diffusion trajectory for the next monomer.
The denoiser is designed around two conditioning axes: a normalized timestep signal that marks the current denoising stage, and a context signal that anchors each refinement step to the prefix chemistry.

The noisy embedding $\mathbf{z}_{t,k}$ is first projected to the model dimension and combined with a sinusoidal timestep embedding $\mathbf{e}_k=\mathrm{MLP}_{t}(\mathrm{SinEmb}(k/K))$.
The resulting state is processed by $L_{\mathrm{den}}$ denoiser blocks:
\begin{equation}
\begin{aligned}
\mathbf{u}^{(0)} &= \mathbf{W}_{z}\mathbf{z}_{t,k} + \mathbf{e}_k,\\
\tilde{\mathbf{u}}^{(l)} &= \mathrm{AdaLN}_{l}\!\left(\mathbf{u}^{(l-1)},\, \mathbf{e}_k\right),\\
\mathbf{u}^{(l)} &= \mathrm{DenoiserBlock}_{l}\!\left(\tilde{\mathbf{u}}^{(l)},\, \mathbf{h}_t\right),\quad l=1,\ldots,L_{\mathrm{den}}.
\end{aligned}
\label{eq:denoiser_block}
\end{equation}
Each denoiser block applies a self-attention-based update to the single noisy-token state, cross-attention to $\mathbf{h}_t$, and a feed-forward transformation with residual connections, layer normalization, and dropout.
AdaLN supplies timestep-dependent scale and shift parameters, whereas cross-attention provides the only path through which the autoregressive history enters the denoising network.

The final block output is normalized and projected back to the Uni-Mol embedding dimension to predict the injected noise:
\begin{equation}
\hat{\boldsymbol{\epsilon}}
= \mathbf{W}_{\mathrm{out}}\mathrm{LN}\!\left(\mathbf{u}^{(L_{\mathrm{den}})}\right)
= \boldsymbol{\epsilon}_\theta(\mathbf{z}_{t,k},\, k,\, \mathbf{h}_t).
\label{eq:noise_prediction}
\end{equation}
The resulting noise estimate is used by the DDPM or DDIM samplers described in Section~\ref{subsubsec:reverse_hybrid_sampling}.

\subsubsection{R-group-aware ring bond predictor}\label{subsec:ring}

To couple macrocyclization with residue generation, PepALD predicts ring-closure bonds within the autoregressive loop.
At each step $t$, the predictor scores whether the newly generated residue $x_t$ should connect to a previous residue $x_j$ ($j<t$) and classifies the bond into one of four R-group link types, $\mathcal{B}=\{\mathrm{R3R3},\mathrm{R1R2},\mathrm{R1R3},\mathrm{R3R2}\}$, which correspond to side chain--side chain linkage, C-terminal backbone--N-terminal backbone linkage, C-terminal backbone--side chain linkage, and N-terminal backbone--side chain linkage, respectively. For each monomer candidate pair $(x_j,x_t)$, the model predicts the ring-closing bond based on hidden features $\mathbf{h}_j$ and $\mathbf{h}_t$, along with their attachment site atom embeddings $\mathbf{E}_{x_t}^{\mathrm{R}}$ and $\mathbf{E}_{x_j}^{\mathrm{R}}$ from Uni-Mol:
\begin{equation}
\left(s_{j,t}^{\mathrm{pos}},\mathbf{l}_{j,t}^{\mathrm{type}}\right)
=
f_{\mathrm{ring}}\!\left(
\mathbf{h}_t,\mathbf{h}_j,
\mathbf{E}_{x_t}^{\mathrm{R}},\mathbf{E}_{x_j}^{\mathrm{R}}
\right),
\quad j<t,
\label{eq:ring_predictor_summary}
\end{equation}
where $s_{j,t}^{\mathrm{pos}}$ is the bond-position score and $\mathbf{l}_{j,t}^{\mathrm{type}}$ contains the logits over R-group link types.
Full definitions of the pair features and prediction heads are provided in Supplementary Section~6.

\subsection{Supervised pre-training and fine-tuning}\label{subsec:training}

PepALD is trained with supervised objectives that target complementary aspects of peptide sequence generation.
The first stage pre-trains the context encoder and denoiser using denoising and auxiliary token-prediction losses.
The second stage fine-tunes the same generator on curated macrocyclic peptides and adds an autoregressive ring-bond objective for learning cyclization topology.

For a $b$-th sample from a mini-batch, let $\Omega$ denote the set of valid non-padding residue positions and let $\mathbf{z}_{b,t,0}=\tilde{\mathbf{e}}_{x_{b,t}}$ be the clean Uni-Mol embedding of token $x_{b,t}$.
For each $(b,t)\in\Omega$, a diffusion step $k$ and Gaussian noise $\boldsymbol{\epsilon}_{b,t}$ are sampled as in Section~\ref{subsec:ald_generative_model}.
The denoiser is optimized by the standard $\epsilon$-prediction objective \citep{ref_ddpm}:
\begin{equation}
\mathcal{L}_{\mathrm{diff}}
=\frac{1}{|\Omega|d}\sum_{(b,t)\in\Omega}
\left\|
\boldsymbol{\epsilon}_{\theta}(\mathbf{z}_{b,t,k}, k, \mathbf{h}_{b,t})
-\boldsymbol{\epsilon}_{b,t}
\right\|_2^2 ,
\label{eq:diff_loss}
\end{equation}
where $d$ is the Uni-Mol embedding dimension and $\mathbf{h}_{b,t}$ is the shifted context vector used to predict the monomer at position $t$.
In parallel, the language-model head receives the same context and predicts the ground-truth monomer identity:
\begin{equation}
\begin{aligned}
p_{\mathrm{LM}}(m\mid \mathbf{h}_{b,t})
&=\left[\operatorname{softmax}(\mathbf{W}_{\mathrm{LM}}\mathbf{h}_{b,t})\right]_m,\\
\mathcal{L}_{\mathrm{CE}}
&=-\frac{1}{|\Omega|}\sum_{(b,t)\in\Omega}
\log p_{\mathrm{LM}}(x_{b,t}\mid \mathbf{h}_{b,t}).
\end{aligned}
\label{eq:ce_loss}
\end{equation}
This auxiliary term regularizes the contextual representation used by the hybrid decoder in Section~\ref{subsubsec:reverse_hybrid_sampling}.

During macrocyclic fine-tuning, the ring predictor is trained on all valid autoregressive candidate pairs.
Let $\mathcal{P}=\{(b,j,t): (b,t)\in\Omega,\; j<t\}$ and let $y_{b,j,t}^{\mathrm{pos}}\in\{0,1\}$ indicate whether residue $x_{b,t}$ forms a ring bond with a previous residue $x_{b,j}$.
The bond-position objective is a weighted binary cross-entropy over the raw logits $s_{b,j,t}^{\mathrm{pos}}$:
\begin{equation}
\begin{aligned}
\pi_{b,j,t} &= \sigma(s_{b,j,t}^{\mathrm{pos}}),\\
\mathcal{L}_{\mathrm{pos}}
&=-\frac{1}{|\mathcal{P}|}
\sum_{(b,j,t)\in\mathcal{P}}
\Big[
w^{+}y_{b,j,t}^{\mathrm{pos}}\log \pi_{b,j,t}\\
&\qquad\qquad\qquad
+(1-y_{b,j,t}^{\mathrm{pos}})\log (1-\pi_{b,j,t})
\Big],
\end{aligned}
\label{eq:pos_loss}
\end{equation}
where $w^{+}=\min(n^{-}/n^{+},50)$ is computed from the positive and negative candidate counts in the mini-batch to compensate for sparse ring closures.
For positive candidate pairs $\mathcal{P}^{+}=\{(b,j,t)\in\mathcal{P}:y_{b,j,t}^{\mathrm{pos}}=1\}$, a separate cross-entropy term classifies the R-group link type:
\begin{equation}
\mathcal{L}_{\mathrm{type}}
=-\frac{1}{|\mathcal{P}^{+}|}
\sum_{(b,j,t)\in\mathcal{P}^{+}}
\log
\frac{\exp(l_{b,j,t,c^{*}}^{\mathrm{type}})}
{\sum_{c\in\mathcal{B}}\exp(l_{b,j,t,c}^{\mathrm{type}})} ,
\label{eq:type_loss}
\end{equation}
where $c^{*}$ is the ground-truth bond type and $\mathcal{B}$ is the four-class R-group bond vocabulary defined in Section~\ref{subsec:ring}.

The supervised objective combines the denoising, token-prediction, and ring-bond terms:
\begin{equation}
\mathcal{L}_{\mathrm{sup}}
=\mathcal{L}_{\mathrm{diff}}
+\lambda_{\mathrm{CE}}\mathcal{L}_{\mathrm{CE}}
+\lambda_{\mathrm{ring}}\left(\mathcal{L}_{\mathrm{pos}}+\mathcal{L}_{\mathrm{type}}\right).
\label{eq:sup_loss}
\end{equation}
During pre-training, the ring objective is disabled; during macrocyclic fine-tuning, all three supervised components are active with configurable weights.

\subsection{Optimization for cell-penetrating peptide design}\label{subsec:preference_optimization}

The downstream task is to optimize PepALD for cyclic cell-penetrating peptide design.
We first construct a permeability-enriched prior generator and then align it with a reward-ranked preference objective.

\subsubsection{Permeability prior fine-tuning}\label{subsubsec:permeability_prior}

For permeability-oriented prior construction, we rank the merged ChEMBL32--CycPeptMPDB prior table by predicted permeability and select the top 1,000 entries for an additional cyclic-only fine-tuning stage.
The resulting permeability-enriched prior model serves as the initialization for the trainable preference-aligned policy.

\subsubsection{Composite reward and WP-DPO objective}\label{subsubsec:dpo_loss}

In the target-specific DPO experiments, candidate rewards are computed from Uni-Dock/Vina docking scores \citep{ref_unidock}.
Because lower docking scores indicate stronger predicted binding, the docking-derived reward is normalized by a robust transformation over the valid candidate set before preference-pair construction.
The resulting scalar score is denoted $R(\mathbf{x}_{1:T})$.
Further details on the sign convention, robust normalization, and invalid-score handling are provided in Supplementary Sections~4 and~7.

Direct preference optimization (DPO) has been used to align generative models from pairwise preferences \citep{ref_dpo}, and its diffusion variant replaces sequence likelihoods with denoising errors along the diffusion process \citep{ref_diffusion_dpo}.
Given a preference pair $(\mathbf{x}_{1:T}^{w},\mathbf{x}_{1:T}^{l})$ with $R(\mathbf{x}_{1:T}^{w})>R(\mathbf{x}_{1:T}^{l})$, we sample a diffusion step $k$ and noise $\boldsymbol{\epsilon}$.
For a model $\theta$, $\overline{\mathrm{MSE}}_{\theta}(\mathbf{x}_{1:T})$ denotes the mean squared difference between the predicted and injected noise, averaged over all valid residue positions of peptide $\mathbf{x}$.
Let
\begin{equation}
\Delta_{\theta}(\mathbf{x}_{1:T})
=\overline{\mathrm{MSE}}_{\mathrm{ref}}(\mathbf{x}_{1:T})
-\overline{\mathrm{MSE}}_{\theta}(\mathbf{x}_{1:T})
\label{eq:dpo_delta}
\end{equation}
be the improvement of the current model over the reference model.
The preference loss is
\begin{equation}
\mathcal{L}_{\mathrm{DPO}}
=-\frac{1}{B}\sum_{b=1}^{B}
\log\sigma\!\left(
\beta_{\mathrm{DPO}}
\left[
\Delta_{\theta}(\mathbf{x}_{1:T}^{w})_{b}
-\Delta_{\theta}(\mathbf{x}_{1:T}^{l})_{b}
\right]\right),
\label{eq:dpo_loss}
\end{equation}
where $B$ is the number of preference pairs in a mini-batch and $\beta_{\mathrm{DPO}}$ controls the strength of the preference margin.
For each pair, the winner and loser are evaluated with the same sampled $k$ and $\boldsymbol{\epsilon}$.
Motivated by DPO-Positive (DPOP), which addresses degradation of preferred samples under standard DPO \citep{ref_dpop}, we add an external winner-protection regularizer to prevent the policy from improving the relative margin by degrading the denoising quality of winner samples:
\begin{equation}
\mathcal{L}_{\mathrm{win}}
=\frac{1}{B}\sum_{b=1}^{B}
\max\left(
0,\,
\overline{\mathrm{MSE}}_{\theta}(\mathbf{x}_{1:T}^{w})_{b}
-
\overline{\mathrm{MSE}}_{\mathrm{ref}}(\mathbf{x}_{1:T}^{w})_{b}
\right).
\label{eq:winner_reg}
\end{equation}
The final WP-DPO objective is
\begin{equation}
\mathcal{L}_{\mathrm{align}}
=\mathcal{L}_{\mathrm{DPO}}
+\alpha_{\mathrm{win}}\mathcal{L}_{\mathrm{win}},
\label{eq:dpo_total_loss}
\end{equation}
where $\alpha_{\mathrm{win}}$ controls the strength of winner protection.

\subsubsection{Iterative candidate-pool construction and preference pairing}\label{subsubsec:pairing}

Preference construction is performed iteratively on model-generated candidate pools rather than on the original training corpus.
At each WP-DPO round, candidates are sampled from the preceding PepALD checkpoint, deduplicated, filtered for valid Uni-Dock/Vina scoring, and ranked by the target-specific reward described above.
Preference pairs are then constructed with a stage-dependent, diversity-aware strategy.
In early WP-DPO rounds, high- and low-reward regions define winner and loser candidate pools, and diversity-aware selection is applied within each pool before pairing.
In later refinement rounds, after the generator has begun to produce competitive binders, we make the preference task more stringent by replacing obvious low-quality losers with hard-negative candidates whose Vina scores are competitive but still less favorable than those of the selected winners.
We then apply diversity-aware reranking to both winner and loser pools using cyclic-residue bigram features, and pair each winner with a structurally similar loser \citep{ref_mmr,ref_drugblip}.
Pairs are retained only when the loser remains significantly worse than its matched winner according to the combined reward.
The final $(\mathbf{x}_{1:T}^{w},\mathbf{x}_{1:T}^{l})$ pairs are fixed before training and reused unchanged during WP-DPO optimization.
Details of diversity selection, hard-negative pairing, and candidate-pool settings are provided in Supplementary Sections~5 and~9.

\subsection{Datasets}\label{subsec:datasets}

We built the training corpus from public peptide resources that provide residue-level monomer annotations and explicit inter-monomer connections.
The main sequence source was ChEMBL32 \citep{ref_chembl}, from which 20,783 HELM sequences were extracted for pre-training.

Then, we performed macrocyclic peptide fine-tuning using \linebreak[4]CycPeptMPDB \citep{ref_cycpeptmpdb}, a curated database of cyclic peptides with membrane-permeability measurements and structural representations.
The raw CycPeptMPDB HELM strings were normalized to a single peptide-polymer identifier, deduplicated, and used as the macrocyclic fine-tuning corpus, yielding 7,348 unique cyclic peptide HELM sequences.
The resulting CycPeptMPDB corpus primarily contains head-to-tail R1--R2 macrocycles, together with R3-mediated terminal-to-internal cyclization patterns.

The monomer vocabulary was compiled from the curated HELM monomer library used in this study, covering monomers observed in ChEMBL32 and CycPeptMPDB together with additional peptide building blocks.
The library contains 3,104 peptide monomer tokens with available R-group attachment sites.
For every monomer, a Uni-Mol representation was precomputed from its molecular structure, including a molecule-level vector and attachment-site vectors for the available R-groups.

\end{methods}

\section{Results}\label{sec:results}

\subsection{Generation quality of prior models}\label{subsec:prior_generation_quality}

We first evaluated the basic generation quality of two PepALD prior models: PepALD$_{\mathrm{ChEMBL}}$, obtained by pretraining on ChEMBL32 HELM peptide sequences, and PepALD$_{\mathrm{CycPeptMPDB}}$, obtained by further fine-tuning PepALD$_{\mathrm{ChEMBL}}$ on CycPeptMPDB cyclic peptide sequences.
We compared these two PepALD priors with HELM-GPT and PepMDLM.
HELM-GPT provides a closely matched HELM-level baseline, as it was likewise pretrained on the ChEMBL database and further fine-tuned on the CycPeptMPDB database \citep{ref_helm_gpt}.
PepMDLM denotes the SMILES-based peptide diffusion prior underlying PepTune before multi-objective inference-time guidance; this prior was trained on a peptide SMILES corpus compiled from CycPeptMPDB, SmProt, and a CycloPs-generated modified-peptide library \citep{ref_peptune}.

Five generation-quality metrics are adapted from MOSES \citep{ref_moses} to evaluate generation quality: validity, uniqueness, novelty, internal diversity, and similarity to the nearest training-set neighbor (SNN). Detailed definitions and implementation details are provided in Supplementary Section~8. Table~\ref{tab:prior_models} presents the evaluation results of the PepALD prior models and reference baselines.
Benefiting from HELM-based monomer-level assembly, a chemically informed foundation-model prior, and autoregressive diffusion for chain-like molecular generation, PepALD$_{\mathrm{ChEMBL}}$ achieved a validity of 0.951 for generating chemically valid peptide structures, substantially higher than HELM-GPT$_{\mathrm{ChEMBL}}$ (0.708) and PepMDLM (0.334).
Beyond validity, consistent with the generative capability of the diffusion-based paradigm, PepALD$_{\mathrm{ChEMBL}}$ and PepMDLM both achieved peak novelty and uniqueness, with values of 1.000.
PepALD$_{\mathrm{ChEMBL}}$ also achieved the highest internal diversity among the evaluated prior models (0.790), outperforming HELM-GPT$_{\mathrm{ChEMBL}}$ (0.740) and PepMDLM (0.709), while maintaining the lowest SNN score (0.441).
After fine-tuning on the CycPeptMPDB dataset, PepALD$_{\mathrm{CycPeptMPDB}}$ continued to show leading performance across these generation-quality metrics.

\begin{figure*}[!t]
\centering
\includegraphics[width=\textwidth]{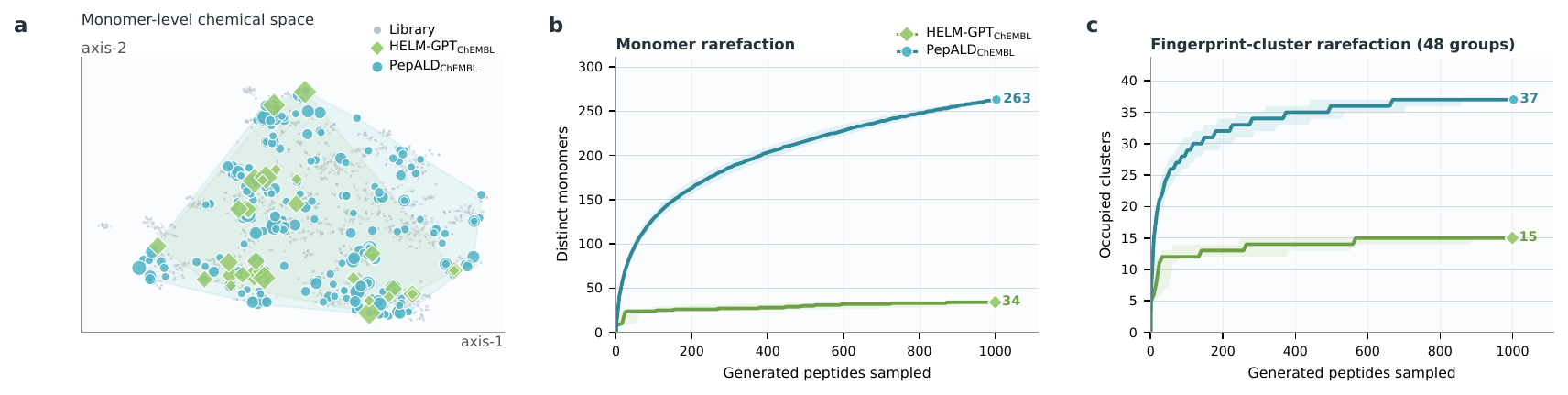}
\caption{Comparison of monomer chemical-space coverage between the PepALD and HELM-GPT prior models before downstream optimization.
(a) Monomer-level chemical-space coverage, where the gray background denotes the full monomer library projected from molecular fingerprints and highlighted points denote monomers used by HELM-GPT and PepALD samples; point size is proportional to monomer usage frequency.
(b) Monomer rarefaction curves over generated peptides.
(c) Fingerprint-cluster rarefaction curves over generated peptides; fingerprint clusters were obtained by applying K-means clustering to SVD-reduced Morgan-fingerprint representations of the full mapped monomer library.
Only clean monomers mapped to the monomer SMILES library are included.}
\label{fig:prior_monomer_coverage}
\end{figure*}

Because PepALD performs diffusion-based generation in a latent space informed by molecular chemical properties, it is not constrained by local minima arising from fixed combinations of building blocks, enabling the generation of diverse peptide molecules.
We therefore examined whether its generalizability was accompanied by broader use of the underlying monomer library.
In the 1,000-sample monomer-level evaluation, PepALD$_{\mathrm{ChEMBL}}$ used 263 clean monomers mapped to the monomer SMILES library, whereas HELM-GPT$_{\mathrm{ChEMBL}}$ used 34 mapped monomers (Fig.~\ref{fig:prior_monomer_coverage}a).
To assess whether this broader monomer usage also covered distinct residue families, we grouped the mapped monomer library into 48 fingerprint-derived clusters.
The rarefaction curves show that this advantage was sustained across sampling depths at both resolutions: PepALD accumulated new monomers more rapidly (Fig.~\ref{fig:prior_monomer_coverage}b) and occupied 37 clusters, compared with 15 clusters for HELM-GPT (Fig.~\ref{fig:prior_monomer_coverage}c).
Together, the monomer- and monomer-cluster-level analyses indicate that PepALD samples more broadly within the existing monomer chemical space, covering more residue identities and fingerprint-derived residue families.

\begin{table*}[!t]
\centering
\caption{Evaluation of PepALD prior models and reference baselines.}
\label{tab:prior_models}
\small
\setlength{\tabcolsep}{0pt}
\begin{tabular*}{\textwidth}{@{\extracolsep{\fill}}lccccc@{}}
\toprule
Model & Validity & Uniqueness & Diversity & SNN & Novelty \\
\midrule
Baseline$_{\mathrm{ChEMBL}}$ & \textbf{1.000} & \textbf{1.000} & \underline{0.768} & 1.000 & 0.000 \\
HELM-GPT$_{\mathrm{ChEMBL}}$ & 0.708 & 0.890 & 0.740 & 0.750 & \underline{0.889} \\
HELM-GPT$_{\mathrm{CycPeptMPDB}}$ & 0.839 & \underline{0.913} & 0.595 & 0.975 & 0.461 \\
PepMDLM & 0.334 & \textbf{1.000} & 0.709 & 0.590 & \textbf{1.000} \\
PepALD$_{\mathrm{ChEMBL}}$ & \underline{0.951} & \textbf{1.000} & \textbf{0.790} & \textbf{0.441} & \textbf{1.000} \\
PepALD$_{\mathrm{CycPeptMPDB}}$ & 0.933 & \textbf{1.000} & 0.726 & \underline{0.504} & \textbf{1.000} \\
\bottomrule
\end{tabular*}
\begin{flushleft}
\footnotesize Baseline$_{\mathrm{ChEMBL}}$ denotes samples from the ChEMBL dataset. PepALD$_{\mathrm{ChEMBL}}$ and PepALD$_{\mathrm{CycPeptMPDB}}$ are the PepALD prior models, corresponding to the ChEMBL-pretrained and CycPeptMPDB-fine-tuned generative models, respectively. HELM-GPT$_{\mathrm{ChEMBL}}$ and HELM-GPT$_{\mathrm{CycPeptMPDB}}$ are the corresponding HELM-GPT prior models; PepMDLM denotes the pretrained PepTune/PepMDLM reference model.
\end{flushleft}
\end{table*}

\subsection{Cell permeability optimization}\label{subsec:permeability_optimization}

To establish a permeability-enriched starting point for subsequent receptor-specific optimization, we next evaluated whether PepALD could be shifted toward cell-permeable macrocyclic peptides while retaining chemical diversity.
As described in Section~\ref{subsubsec:permeability_prior}, PepALD$_{\mathrm{perm}}$ was initialized from PepALD$_{\mathrm{CycPeptMPDB}}$ and further fine-tuned on the top 1,000 training-set molecules with the highest predicted permeability scores.

For permeability scoring, we adopted the random-forest predictor previously used in HELM-GPT \citep{ref_helm_gpt}, which achieved a Spearman correlation of 0.82 on a held-out macrocyclic-peptide test set.
Accordingly, we used this predictor to compute the average predicted permeability (less negative values indicating better membrane permeability) of cyclic peptides generated by each method.
Across training stages, the predicted permeability scores progressively improved (Supplementary Fig.~S2).

\begin{wrapfigure}{r}{0.43\textwidth}
\vspace{-8pt}
\centering
\includegraphics[width=\linewidth,trim=0 3pt 0 2pt,clip]{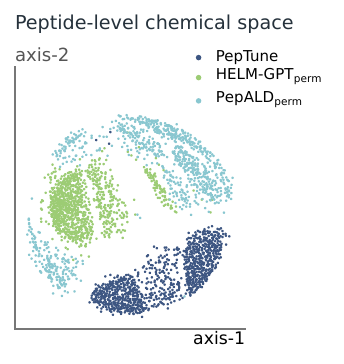}
\vspace{-8pt}
\caption{Peptide-level chemical-space coverage after permeability-oriented generation, shown as an MDS projection of pairwise molecular-fingerprint distances for valid samples from PepTune, HELM-GPT$_{\mathrm{perm}}$, and PepALD$_{\mathrm{perm}}$.}
\label{fig:permeability_peptide_space}
\vspace{-10pt}
\end{wrapfigure}

\begin{figure}[H]
\begingroup
\setlength{\abovecaptionskip}{4pt plus 1pt minus 1pt}
\setlength{\belowcaptionskip}{4pt plus 1pt minus 1pt}
\centering
\includegraphics[width=0.92\textwidth]{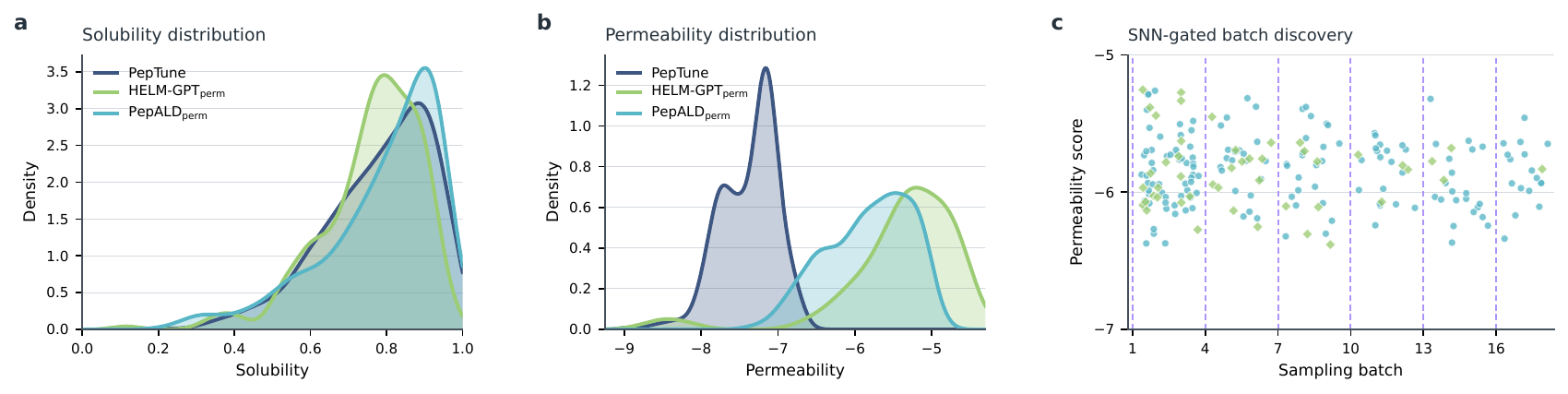}
\caption{Solubility and permeability evaluation of PepALD$_{\mathrm{perm}}$, HELM-GPT$_{\mathrm{perm}}$, and PepTune under permeability-oriented optimization.
(a) Predicted soluble-class probability distributions for valid samples from PepTune, HELM-GPT$_{\mathrm{perm}}$, and PepALD$_{\mathrm{perm}}$.
(b) Predicted permeability-score distributions of valid generated samples from the same generators; scores $\leq -10$ were excluded.
(c) SNN-gated batch discovery from equal-size raw-sample batches.
Each batch drew 40 samples per model, and the remaining candidates were retained only if their maximum Morgan-fingerprint Tanimoto similarity to the previously accepted pool was below 0.60.
For readability, panel (c) displays retained candidates with permeability scores between $-7$ and $-5$.}
\label{fig:permeability_optimization}
\endgroup
\end{figure}

Table~\ref{tab:permeability_models} compares PepALD$_{\mathrm{perm}}$ with permeability-oriented baselines.
HELM-GPT$_{\mathrm{perm}}$ denotes the HELM-GPT prior optimized for permeability, whereas PepTune represents the PepMDLM model using permeability-oriented guidance.
Because adequate aqueous solubility is needed to maintain the concentration gradient that drives membrane flux, we also report the predicted soluble-class probability alongside permeability \citep{ref_peptune}, with the valid-sample solubility distributions shown in Fig.~\ref{fig:permeability_optimization}a.
In terms of permeability, PepALD$_{\mathrm{perm}}$ achieved a mean predicted permeability of -5.8215, marginally inferior to HELM-GPT$_{\mathrm{perm}}$ (-5.3964) among the generative models and substantially better than PepTune (-7.3783), as illustrated by the predicted permeability distributions in Fig.~\ref{fig:permeability_optimization}b.
However, the higher permeability of HELM-GPT$_{\mathrm{perm}}$ was accompanied by a marked reduction in generation usability.
Its validity was only 0.253, compared with 0.955 for PepALD$_{\mathrm{perm}}$ and 0.341 for PepTune.
PepALD$_{\mathrm{perm}}$ also retained stronger generation-quality characteristics, with an internal diversity of 0.660 that exceeded both HELM-GPT$_{\mathrm{perm}}$ and PepTune, together with a moderate SNN of 0.585.
Consistent with this diversity profile, the permeability-oriented PepALD$_{\mathrm{perm}}$ samples occupied a broad peptide-level chemical-space region while retaining partial overlap with HELM-GPT$_{\mathrm{perm}}$ (Fig.~\ref{fig:permeability_peptide_space}), indicating that the permeability shift did not collapse generation into a narrow neighborhood.

\medskip

\begingroup
\linespread{1.06}\selectfont
\setlength{\parskip}{4pt plus 1pt minus 1pt}
The importance of validity and diversity was further supported by the SNN-gated batch-discovery analysis in Fig.~\ref{fig:permeability_optimization}c.
When equal-size raw-sample batches were filtered by an SNN $<0.60$ novelty gate, HELM-GPT$_{\mathrm{perm}}$ produced progressively fewer retained candidates during repeated sampling, suggesting that its generated samples increasingly revisited chemically similar neighborhoods.
In contrast, PepALD$_{\mathrm{perm}}$ continued to provide nonredundant candidates within the desired permeability range across sampling batches.

Through a permeability-oriented strategy, PepALD$_{\mathrm{perm}}$ not only exhibits superior properties in terms of permeability scores, but also preserves the structural plausibility, diversity, and solubility of the generated cyclic peptides, which suggests that PepALD$_{\mathrm{perm}}$ provides an effective and balanced framework for the in silico design of cyclic peptides.
\par
\endgroup

\par\smallskip
\noindent\begin{minipage}{\textwidth}
\centering
\small
\captionof{table}{Evaluation of permeability-oriented generation.}
\label{tab:permeability_models}
\setlength{\tabcolsep}{0pt}
\begin{tabular*}{\textwidth}{@{\extracolsep{\fill}}lccccc@{}}
\toprule
Model & Permeability & Solubility & Diversity & SNN & Validity \\
\midrule
Baseline$_{\mathrm{ChEMBL}}$ & -7.104 & \underline{0.837} & \underline{0.768} & 1.000 & \textbf{1.000} \\
HELM-GPT$_{\mathrm{perm}}$ & \textbf{-5.396} & 0.759 & 0.553 & 0.621 & 0.253 \\
PepTune & -7.378 & 0.825 & 0.611 & 0.537 & 0.341 \\
PepALD$_{\mathrm{perm}}$ & \underline{-5.822} & 0.788 & 0.660 & 0.585 & \underline{0.955} \\
PepALD$_{\mathrm{ChEMBL}}$ & -7.069 & 0.821 & \textbf{0.790} & \textbf{0.441} & 0.951 \\
PepALD$_{\mathrm{CycPeptMPDB}}$ & -6.669 & \textbf{0.840} & 0.726 & \underline{0.504} & 0.933 \\
\bottomrule
\end{tabular*}
\begin{flushleft}
\footnotesize Baseline$_{\mathrm{ChEMBL}}$ denotes samples from the ChEMBL dataset. PepALD$_{\mathrm{ChEMBL}}$ and PepALD$_{\mathrm{CycPeptMPDB}}$ are the PepALD priors before permeability fine-tuning. HELM-GPT$_{\mathrm{perm}}$ is the permeability-oriented HELM-GPT baseline; PepTune denotes the PepTune/PepMDLM reference model. Permeability is higher-is-better, and solubility is the PepTune predicted soluble-class probability.
\end{flushleft}
\end{minipage}
\par\smallskip

\FloatBarrier
\subsection{Optimization for specific tasks}\label{subsec:specific_task_optimization}

We further applied PepALD to two target-specific macrocyclic peptide optimization cases: SPRY domain-containing SOCS box protein 2 (SPSB2, PDB ID: 6DN5) and Chorismate Mutase from Mycobacterium tuberculosis (MtbCM, PDB ID: 9BT3).
To optimize and evaluate the ability of generated macrocyclic peptides to bind the target proteins, we utilized docking scores from Uni-Dock, a GPU-accelerated Vina-style docking engine, as both the candidate rewards and the affinity metric \citep{ref_unidock}.
To more comprehensively evaluate the applicability of PepALD in macrocyclic peptide drug-design scenarios, the predicted membrane permeability and aqueous solubility of the generated peptides were also incorporated as evaluation metrics.

\subsubsection{Cyclic peptide generation for SPSB2}\label{subsubsec:spsb2_generation}

SPSB2 is a SPRY domain-containing adaptor protein that recruits iNOS for ubiquitin-mediated degradation, thereby suppressing nitric oxide production and modulating host antimicrobial responses.
Designing cyclic peptide inhibitors against the SPSB2-iNOS interface can potently and selectively stabilize iNOS levels, offering a structure-guided strategy for enhancing nitric oxide-dependent clearance of intracellular pathogens.

In this task, we aim to generate cell-permeable macrocyclic peptide binders targeting SPSB2 (PDB ID: 6DN5) using PepALD$_{\mathrm{perm}}$.
Given the exploratory capacity of PepALD, we did not introduce any additional 6DN5-specific peptide or ligand set as prior information.
Instead, the WP-DPO policy was initialized directly from PepALD$_{\mathrm{perm}}$ and optimized using the strategy described in Section~\ref{subsec:preference_optimization}.

The Vina-score distributions in Fig.~\ref{fig:specific_task_optimization}a show a steady shift toward more favorable docking scores during WP-DPO.
Through ten optimization rounds, PepALD$_{\mathrm{SPSB2}}$ can generate macrocyclic peptides with more favorable docking scores, improving the mean Vina score from -5.832 for PepALD$_{\mathrm{perm}}$ to -7.020 for PepALD$_{\mathrm{SPSB2}}$.
The top-ranked subsets exceeded the reference ligand in predicted docking affinity, with mean Vina scores of -8.037 for the top $10\%$ and -7.746 for the top $25\%$, compared with -7.627 for the reference ligand.
At the same time, the optimized population retained a higher mean predicted solubility than the reference ligand (0.863 versus 0.785) and maintained internal diversity (0.654), while its predicted permeability remained more favorable (-6.593 versus -7.181).
For representative structural visualization, we selected top-ranked candidates by jointly considering Vina score, predicted permeability, and predicted solubility; two SPSB2-targeted examples are shown docked to 6DN5 in Fig.~\ref{fig:specific_task_optimization}b,c.

\subsubsection{Cyclic peptide generation for MtbCM}\label{subsubsec:mtbcm_generation}

Chorismate mutase from Mycobacterium tuberculosis (Mtb) is a secreted enzyme implicated in bacterial virulence and host-pathogen interactions. 
Designing cyclic peptide inhibitors against Mtb chorismate mutase (MtbCM) has the potential to enable potent enzymatic inhibition and provide new ligand scaffolds for tuberculosis therapies amid rising antibiotic resistance.

In this task, we aim to generate cell-permeable macrocyclic peptide binders targeting MtbCM (PDB ID: 9BT3) using PepALD$_{\mathrm{perm}}$. The WP-DPO optimization procedure followed the same workflow as the SPSB2-binding cyclic peptide generation task described above.

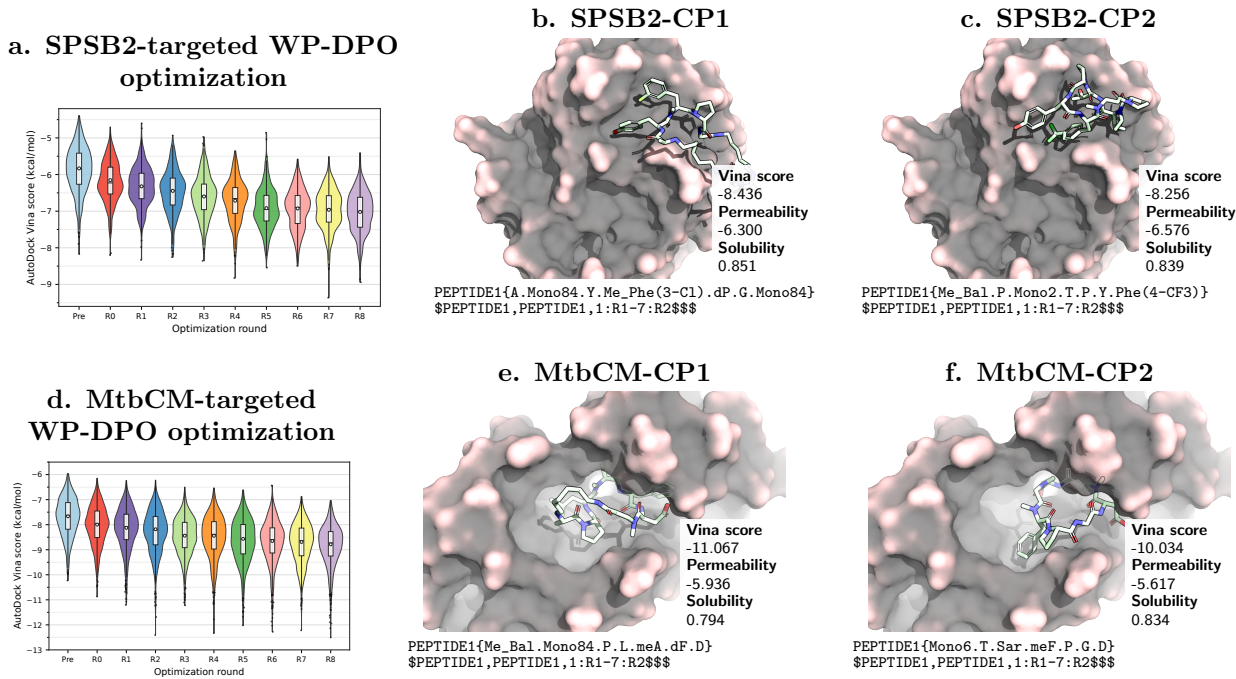
\begin{figure}[H]
\centering
\small
\begingroup
\newcommand{\dockposepanel}[9]{%
\begin{minipage}[t]{0.315\textwidth}
\centering
\textbf{#1. #2}\\[3pt]
\begin{tikzpicture}
\node[inner sep=0pt] (poseimage)
{\includegraphics[width=\linewidth,height=1.25in,keepaspectratio,#4]{#3}};
\node[
anchor=south east,
fill=white,
fill opacity=0.90,
text opacity=1,
inner xsep=2.2pt,
inner ysep=1.6pt,
font=\tiny\sffamily,
align=left
] at ([xshift=-1pt,yshift=1pt]poseimage.south east)
{\textbf{Vina score}\\#5\\[-0.5pt]\textbf{Permeability}\\#6\\[-0.5pt]\textbf{Solubility}\\#7};
\end{tikzpicture}\\[-1pt]
\parbox{\linewidth}{\raggedright\tiny\ttfamily
\detokenize{#8}\\[-1pt]\detokenize{#9}\par}
\end{minipage}%
}
\begin{minipage}[t]{0.315\textwidth}
\centering
\vspace*{3pt}
\textbf{a. SPSB2-targeted WP-DPO optimization}\\[3pt]
\includegraphics[width=\linewidth,height=1.25in,keepaspectratio]{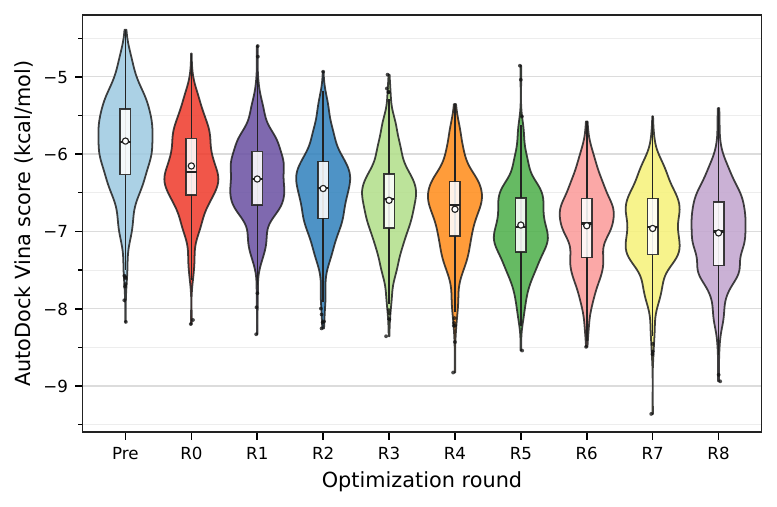}
\end{minipage}\hfill
\dockposepanel{b}{SPSB2-CP1}
{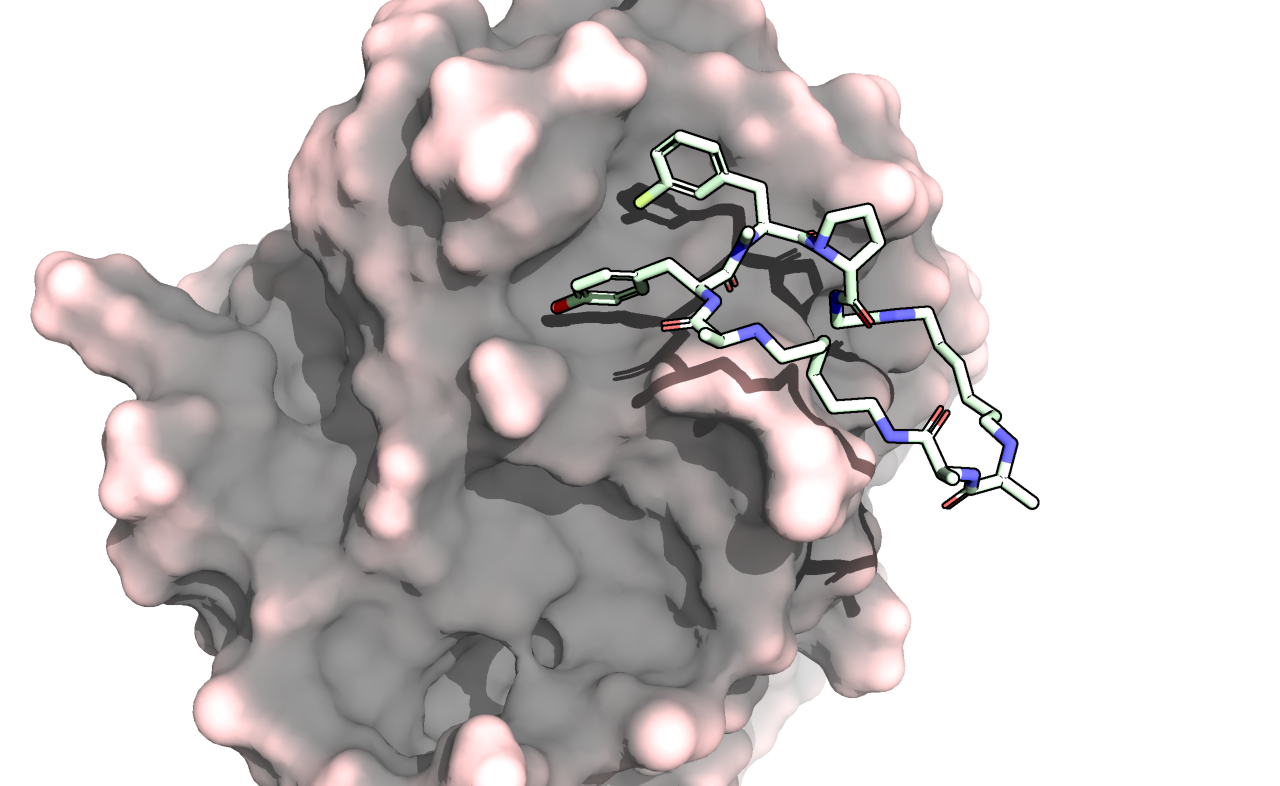}
{trim=0 0 60 0,clip}
{-8.436}{-6.300}{0.851}
{PEPTIDE1{A.Mono84.Y.Me_Phe(3-Cl).dP.G.Mono84}}
{$PEPTIDE1,PEPTIDE1,1:R1-7:R2$$$}\hfill
\dockposepanel{c}{SPSB2-CP2}
{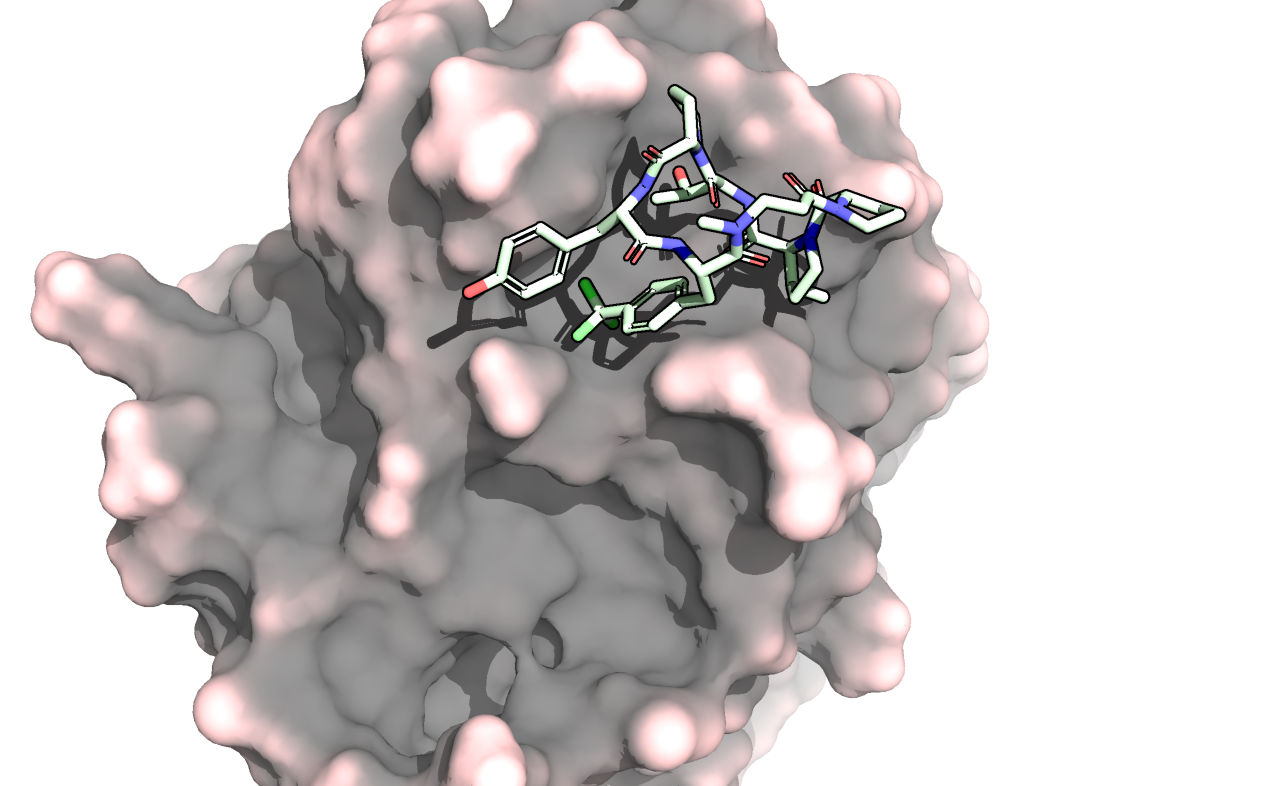}
{trim=0 0 60 0,clip}
{-8.256}{-6.576}{0.839}
{PEPTIDE1{Me_Bal.P.Mono2.T.P.Y.Phe(4-CF3)}}
{$PEPTIDE1,PEPTIDE1,1:R1-7:R2$$$}
\par\vspace{7pt}
\begin{minipage}[t]{0.28\textwidth}
\centering
\vspace*{3pt}
\textbf{d. MtbCM-targeted WP-DPO optimization}\\[3pt]
\includegraphics[width=\linewidth,height=1.25in,keepaspectratio]{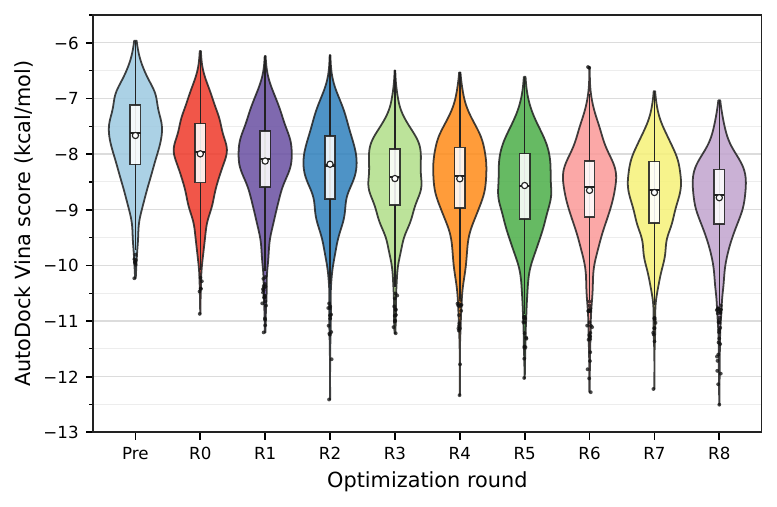}
\end{minipage}\hfill
\dockposepanel{e}{MtbCM-CP1}
{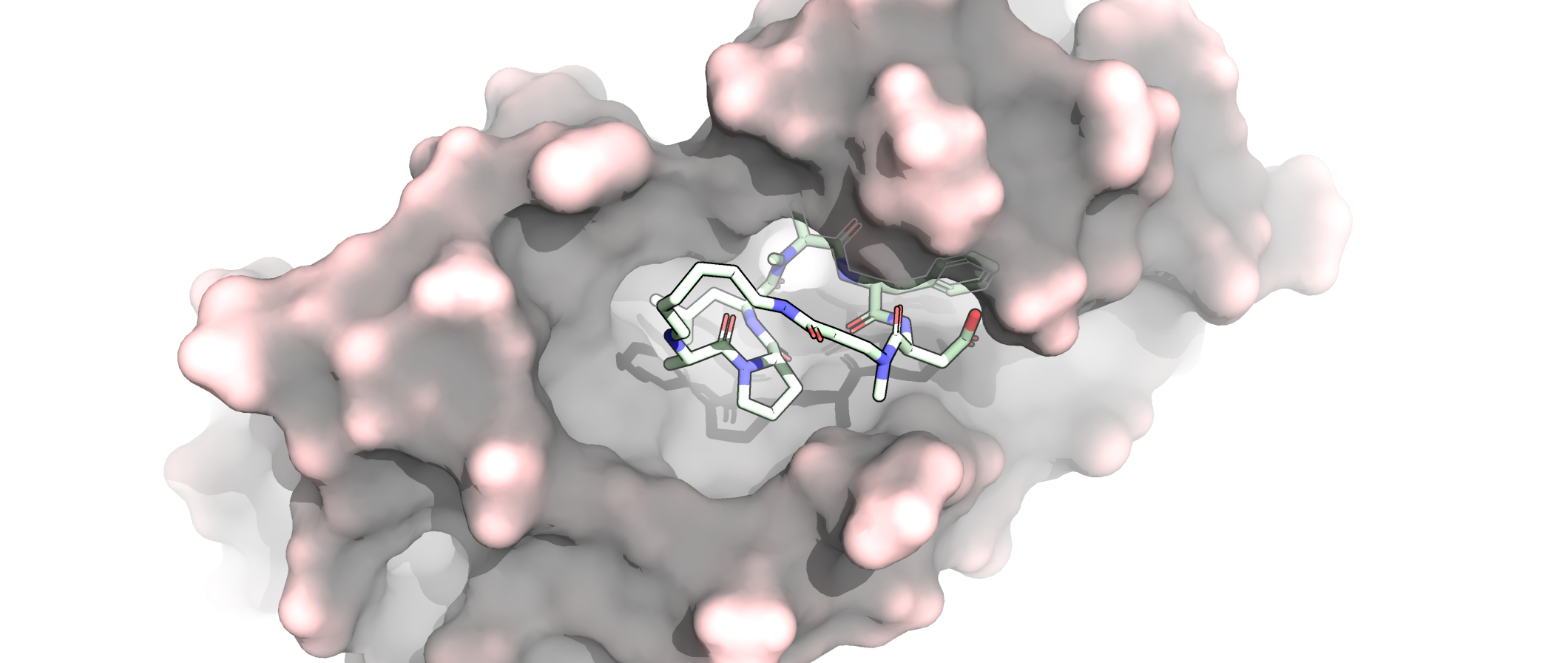}
{trim=410 0 360 0,clip}
{-11.067}{-5.936}{0.794}
{PEPTIDE1{Me_Bal.Mono84.P.L.meA.dF.D}}
{$PEPTIDE1,PEPTIDE1,1:R1-7:R2$$$}\hfill
\dockposepanel{f}{MtbCM-CP2}
{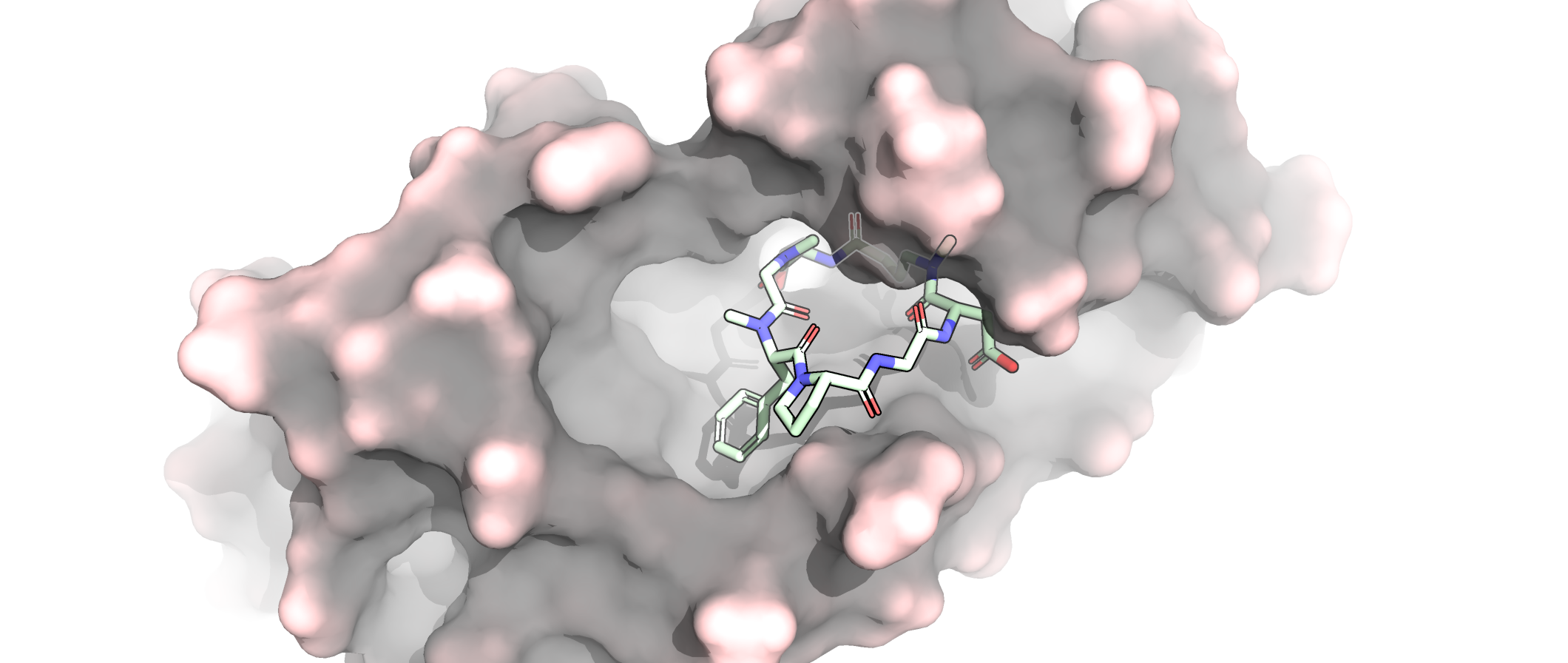}
{trim=410 0 360 0,clip}
{-10.034}{-5.617}{0.834}
{PEPTIDE1{Mono6.T.Sar.meF.P.G.D}}
{$PEPTIDE1,PEPTIDE1,1:R1-7:R2$$$}
\caption{Target-specific WP-DPO optimization redirects PepALD toward receptor-favorable macrocyclic peptides while retaining permeability-related properties.
(a) Vina score distributions for SPSB2-targeted cyclic peptide generation, targeting the SPSB2--iNOS interface represented by 6DN5, from PepALD$_{\mathrm{perm}}$ through iterative WP-DPO rounds.
(b,c) Representative optimized SPSB2 cyclic-peptide candidates docked to 6DN5.
(d) Vina score distributions for MtbCM-targeted cyclic peptide generation, targeting Mtb chorismate mutase (MtbCM) represented by 9BT3, under the same optimization protocol.
(e,f) Representative optimized MtbCM cyclic-peptide candidates docked to 9BT3.}
\label{fig:specific_task_optimization}
\endgroup
\end{figure}

The MtbCM-targeted task showed the same self-improving behavior under generated-data WP-DPO.
The mean Vina score improved from \mbox{-7.641} for PepALD$_{\mathrm{perm}}$ to \mbox{-8.645} for PepALD$_{\mathrm{MtbCM}}$, and the top-ranked generated subsets reached substantially stronger docking scores.
In particular, the top $10\%$ subset achieved a mean Vina score of -10.324, and the top $25\%$ subset reached -9.781, matching the reference ligand-level docking performance (-9.730).
The optimized cyclic peptides exhibited substantially better predicted permeability than the reference ligand, with a mean score of -5.889 compared with -7.089, while maintaining high solubility (0.857) and diversity (0.660).
Using the same selection criterion, two representative MtbCM-targeted candidates with favorable Vina-score, predicted-permeability, and predicted-solubility profiles are shown docked to 9BT3 in Fig.~\ref{fig:specific_task_optimization}e,f.

Taken together, the SPSB2- and MtbCM-targeted optimization tasks indicate that the permeability-enriched PepALD prior can be redirected to new protein structures through WP-DPO-driven optimization using only model-generated preference data, while maintaining favorable permeability and solubility.
This highlights the practical potential of PepALD for cyclic peptide design.

\begin{table}[H]
\centering
\footnotesize
\setlength{\tabcolsep}{0pt}
\begin{minipage}[t]{0.495\textwidth}
\centering
\textbf{(a) PepALD-generated cyclic peptides binding 6DN5}\\[1pt]
\renewcommand{\arraystretch}{1.08}
\begin{tabular*}{\linewidth}{@{\extracolsep{\fill}}l
S[table-format=-1.3]
S[table-format=1.3]
S[table-format=1.3]
S[table-format=-2.3]@{}}
\toprule
Model & \multicolumn{1}{c}{Permeability} & \multicolumn{1}{c}{Solubility} & \multicolumn{1}{c}{Diversity} & \multicolumn{1}{c}{Vina score} \\
\midrule
PepALD$_{\mathrm{perm}}$ & -5.804 & 0.794 & 0.656 & -5.832 \\
PepALD$_{\mathrm{SPSB2}}$ & -6.593 & 0.863 & 0.654 & -7.020 \\
Top $10\%$ & -6.622 & 0.819 & 0.608 & -8.037 \\
Top $25\%$ & -6.619 & 0.836 & 0.634 & -7.746 \\
Reference ligand & -7.181 & 0.785 & \multicolumn{1}{c}{--} & -7.627 \\
\bottomrule
\end{tabular*}
\end{minipage}\hfill%
\begin{minipage}[t]{0.495\textwidth}
\centering
\textbf{(b) PepALD-generated cyclic peptides binding 9BT3}\\[1pt]
\renewcommand{\arraystretch}{1.08}
\begin{tabular*}{\linewidth}{@{\extracolsep{\fill}}l
S[table-format=-1.3]
S[table-format=1.3]
S[table-format=1.3]
S[table-format=-2.3]@{}}
\toprule
Model & \multicolumn{1}{c}{Permeability} & \multicolumn{1}{c}{Solubility} & \multicolumn{1}{c}{Diversity} & \multicolumn{1}{c}{Vina score} \\
\midrule
PepALD$_{\mathrm{perm}}$ & -5.804 & 0.794 & 0.656 & -7.641 \\
PepALD$_{\mathrm{MtbCM}}$ & -5.889 & 0.857 & 0.660 & -8.645 \\
Top $10\%$ & -5.987 & 0.876 & 0.631 & -10.324 \\
Top $25\%$ & -5.934 & 0.865 & 0.631 & -9.781 \\
Reference ligand & -7.089 & 0.917 & \multicolumn{1}{c}{--} & -9.730 \\
\bottomrule
\end{tabular*}
\end{minipage}
\caption{Target-specific optimization metrics.
(a) Optimization of PepALD-generated cyclic peptides binding 6DN5.
(b) Optimization of PepALD-generated cyclic peptides binding 9BT3.
Each panel reports scalar metrics for PepALD$_{\mathrm{perm}}$, the final task policy, top-ranked generated subsets, and the reference ligand.
Permeability, solubility, and diversity are higher-is-better; Vina score is lower-is-better.}
\label{tab:spsb2_task}
\label{tab:mtbcm_task}
\label{tab:specific_task_metrics}
\end{table}

\FloatBarrier

\vspace{-12pt}
\section{Discussion and conclusion}\label{sec:conclusion}

In this study, we developed PepALD, a chemically grounded generative framework for HELM-represented macrocyclic peptides. The model combines residue-level HELM syntax, chemistry-informed monomer representation, autoregressive latent diffusion, R-group-aware ring-bond prediction, and diffusion-adapted preference optimization. This design addresses two limitations that are particularly important for macrocyclic peptide discovery: atom-level molecular strings are long and difficult to optimize in sparse peptide data regimes, whereas purely symbolic HELM language models do not directly expose the underlying chemistry of natural and non-natural monomers. By generating each residue through a context-conditioned diffusion process in the Uni-Mol latent space and then projecting to chemically admissible HELM tokens, PepALD preserves the compactness and topology awareness of HELM while allowing the generator to exploit molecular similarity between monomers.

The empirical results support this formulation. Before target-specific optimization, PepALD outperformed HELM-GPT and PepMDLM in validity, uniqueness, novelty, diversity, and SNN profiles. The monomer-coverage analysis further showed that the gain was not merely a diffuse increase in peptide-level variability: PepALD sampled more diverse monomers than HELM-GPT, indicating broader use of the available building blocks. After permeability-oriented fine-tuning, PepALD shifted toward the high-permeability regime while retaining substantially higher validity and diversity than the permeability-optimized baselines. Finally, in two receptor-specific case studies, WP-DPO-enhanced PepALD improved docking score distributions without requiring target-specific supervised fine-tuning, while maintaining favorable predicted permeability, solubility, and internal diversity. These results suggest that a permeability-enriched PepALD prior can serve as a reusable starting point for macrocyclic peptide generation targeting new proteins.

Despite PepALD's promising performance in generating bioactive cyclic peptides, a key limitation persists in its building block generation paradigm. The current approach generates latent embeddings via diffusion and matches them to the closest pre-existing building blocks in our curated library. While this ensures high synthetic feasibility, it imposes a hard boundary on accessible chemical space, fundamentally constraining peptide diversity to the library's size and composition. To overcome this limitation, we propose extending the diffusion trajectory directly from generated latent embeddings to produce entirely novel building blocks, rather than terminating at the matching step. This next-generation framework would integrate predictive modules for synthetic feasibility and chemical stability into the diffusion network, enabling the model to learn chemical validity and synthetic accessibility during denoising. This paradigm shift would liberate PepALD from pre-defined library constraints, unlocking vastly expanded cyclic peptide chemical space and facilitating the discovery of next-generation peptide therapeutics.

In conclusion, PepALD provides a unified and extensible framework for \textit{de novo} macrocyclic peptide design. By coupling HELM grammar with chemically meaningful monomer embeddings and diffusion-based preference optimization, it generates diverse valid macrocycles and can be redirected toward permeability- and affinity-relevant objectives using reinforcement learning. These properties make PepALD a promising computational platform for exploring non-natural macrocyclic peptide space and for accelerating the discovery of cell-permeable cyclic peptides against challenging intracellular targets. We anticipate that future wet-lab experiments will further validate the ability of PepALD to generate experimentally actionable macrocyclic peptide candidates.

\section*{Authors' contributions}

J.Z. and Z.G. designed the research. J.Z. and Z.G. wrote source code and performed the experiments. Z.G. and W.J. supervised the project. S.Y. guided the experiments. J.Z., Z.G., W.J. and S.Y. analyzed the experimental results. J.Z. wrote the manuscript. Z.G. and W.J. revised the manuscript. All authors read and approved the final manuscript.

\section*{Funding}

This work is supported in part by the National Natural Science Foundation of China under Grants 62306014 and 12501344, the Postdoctoral Fellowship Program (Grade A) of CPSF under Grants BX20250376 and BX20240239, the China Postdoctoral Science Foundation under Grant 2024M762201, and the Sichuan Science and Technology Program under Grants 2025ZNSFSC1506 and 2025ZNSFSC0808.

\section*{Conflict of Interest}

The authors declare that they have no competing interests.

\bibliographystyle{natbib}
\bibliography{bibliography}

\end{document}